\documentclass{article}
\usepackage{spconf,amsmath,graphicx, adjustbox, inputenc}

\title{Camera Calibration through Camera Projection Loss}
\name{Talha Hanif Butt, Murtaza Taj}
\address{Computer Vision and Graphics Lab\\
 Lahore University of Management Sciences, Lahore, Pakistan\\
 thanifbutt@gmail.com,  murtaza.taj@lums.edu.pk\\
 https://cvlab.lums.edu.pk/}
\begin{document}
%\ninept
%
\maketitle
\begin{abstract}
Camera calibration is a necessity in various tasks including 3D reconstruction, hand-eye coordination for a robotic interaction, autonomous driving, etc. In this work we propose a novel method to predict extrinsic (baseline, pitch, and translation), intrinsic (focal length and principal point offset) parameters using an image pair. Unlike existing methods, instead of designing an end-to-end solution, we proposed a new representation that incorporates camera model equations as a neural network in a multi-task learning framework. We estimate the desired parameters via novel \emph{camera projection loss} (CPL) that uses the camera model neural network to reconstruct the 3D points and uses the reconstruction loss to estimate the camera parameters. To the best of our knowledge, ours is the first method to jointly estimate both the intrinsic and extrinsic parameters via a multi-task learning methodology that combines analytical equations in learning framework for the estimation of camera parameters. We also proposed a novel CVGL Camera Calibration dataset using CARLA Simulator~\cite{Dosovitskiy17}. Empirically, we demonstrate that our proposed approach achieves better performance with respect to both deep learning-based and traditional methods on 8 out of 10 parameters evaluated using both synthetic and real data. Our code and generated dataset are available at https://github.com/thanif/Camera-Calibration-through-Camera-Projection-Loss.

\end{abstract}
\section{Introduction}
\label{sec:intro}

Camera calibration deals with finding the five intrinsic ( focal length, image sensor format, and principal point) and six extrinsic (rotation, translation) parameters of the specific camera. %Camera calibration is useful in many computer vision tasks such as image alignment and 3D reconstruction which are building blocks for many important applications including self driving cars, augmented reality, 3D pose estimation.

%The process of image formation is well understood and has been studied extensively in computer vision ~\cite{hartley2003multiple}, allowing for very precise calibration of cameras when there are enough geometric constraints to fit the camera model. Multi-view geometry (MVG) based methods typically fine corresponding points between images to generate enough constraints to solve the camera model equations and thus calculate the calibration parameters. Over the past several years, a number of calibration techniques have been proposed ~\cite{unnikrishnan2005fast,geiger2012automatic,levinson2013automatic,pandey2012automatic,taylor2015motion}. Yet, the vast majority of these techniques depend on specific calibration targets such as checkerboards, and require significant amounts of manual effort~\cite{unnikrishnan2005fast,geiger2012automatic}. The process of image correspondence is often automated via strong cues such as vanishing points and straight lines that can be used to recover the camera parameters~\cite{caprile1990using,deutscher2002automatic}. However,  MVG-based methods lack robustness due to images taken in unstructured environments, with varying illumination conditions.

\begin{figure}
\centering
    {\includegraphics[trim={5cm 12cm 8cm 0},clip,width=0.95\columnwidth]{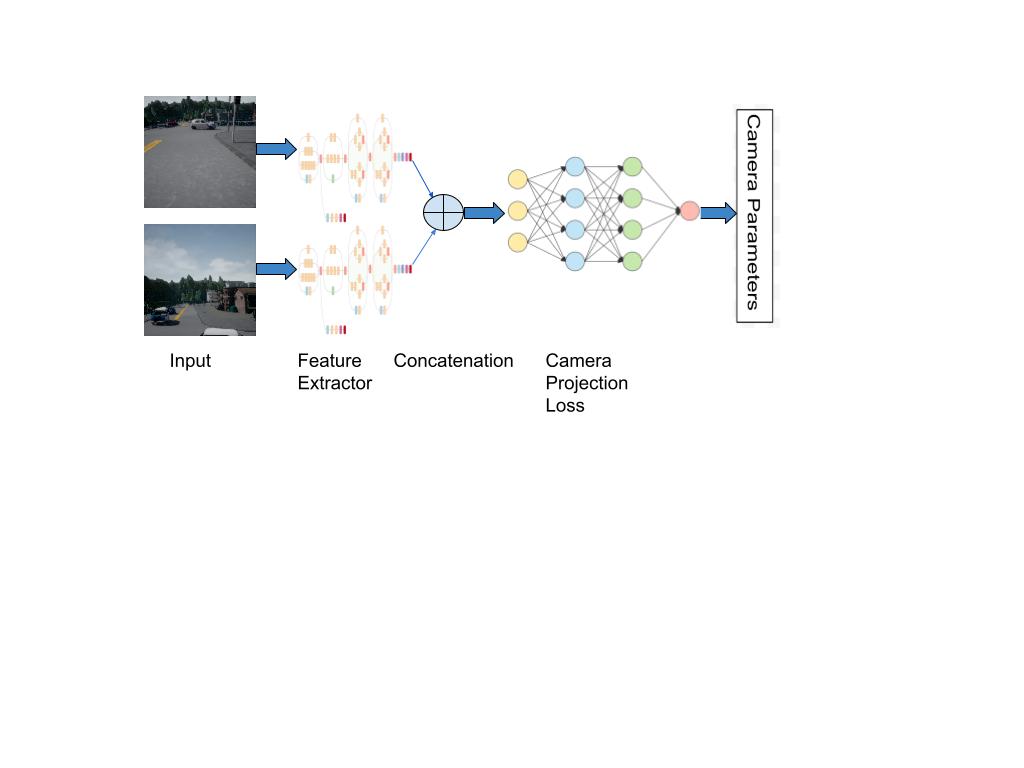}} 
    \caption{Our method estimates extrinsic (baseline, pitch and translation) and intrinsic (focal length and principal point offset) parameters using pre-trained Inception-v3~\cite{szegedy2016rethinking} and the proposed Camera Projection Loss.} %Lambda layers representation of which is shown in %Fig.~\ref{fig:camera_projection_loss}.}
    \label{fig:flowdiagram}
\end{figure}

 Most of the existing methods usually ignore the underlying mathematical formulation of the camera model and instead propose an end-to-end framework to directly estimate the desired parameters~\cite{workman2015deepfocal,rong2016radial,hold2018perceptual, lopez2019deep, zhai2016detecting, detone2016deep,bogdan2018deepcalib,zhang2020deepptz,workman2016horizon,barreto2006unifying}. Thus they are difficult to interpret for real-world applications and have so far been able to mainly estimate the focal length of the camera via single image only~\cite{detone2016deep,bogdan2018deepcalib,zhang2020deepptz}. %In this work we not only estimate $8$ out of $11$ calibration parameters, we also estimate baseline and disparity. Furthermore, we propose a learning-based method that rely on the underlying mathematical equations of pinhole camera model. 
 The major contributions of our work are as follows:

\begin{itemize}
    \item To the best of our knowledge, this work is the first learning-based method to jointly estimate both intrinsic and extrinsic camera parameters including camera baseline, disparity, pitch, translation, focal length and principal point offset.
    \item The existing learning based approaches~\cite{detone2016deep, workman2015deepfocal, zhang2020deepptz} have not been applied to the estimation of all $10$ camera parameters due to lack of any dataset. We addressed this limitation by generating a synthetic dataset from two towns in CARLA~\cite{Dosovitskiy17} simulation consisting of $49$ different camera settings.
    \item Unlike existing methods, instead of designing an end-to-end solution to directly estimate the desired parameters~\cite{detone2016deep}, we proposed a new representation that represents camera model equations as a neural network in a multi-task learning (MTL) framework.
    \item We proposed a novel \emph{camera projection loss} (CPL) that combines analytical equations in learning framework. %We use the proposed camera model neural network to reconstruct 3D point cloud and use the reconstruction loss to estimate the camera parameters. 
\end{itemize}

\section{Proposed Method}
\label{sec:pm}

%\begin{figure*}[t]
%\begin{center}
%\begin{tabular}{ccc}
%    \multicolumn{3}{c}{\includegraphics[width=0.9\textwidth]{images/3.png}}\\
%	\includegraphics[width=0.25\textwidth]{images/xcam.jpg}&
%	\includegraphics[width=0.28\textwidth]{images/ycam.jpg}&
%	\includegraphics[width=0.45\textwidth]{images/zcam.jpg}\\
%	(a) & (b) & (c)\\
%\end{tabular}
%\end{center}
%\caption{Camera Projection Loss (CPL) in the form of Lambda layers. Lambda layers have been used to implement the loss using (Eq.~\ref{eq-4a} - Eq.~\ref{eq-5c}). (a-c) are sub components CPL showing Lambda layer representation of $x_{cam}$, $y_{cam}$ and $z_{cam}$ respectively. This depicts the implementation of Eqs.~\ref{eq-4a}-\ref{eq-4c} as neural network respectively.}
%\label{fig:camera_projection_loss}
%\end{figure*}

%We briefly summarize our method and describe the details in subsequent sections. 
We propose to utilize multi-task learning by incorporating mathematical equations through a new loss function embedded as a neural network for better representation while learning.
We train a convolutional neural network to predict the extrinsic and intrinsic camera parameters. To achieve this, we use dependent regressors that share a common network architecture as the feature extractor. We use a Inception-v3~\cite{szegedy2016rethinking} pretrained on ImageNet~\cite{russakovsky2015imagenet}
as a feature extractor followed by the Lambda layers for loss computation with $13$ regressors, 10 of which correspond to the camera parameters while 3 correspond to the 3D point cloud. Instead of training these regressors to predict the focal length, principal point, baseline, pitch, and translation, we use proxy variables that are not visible in the image and are dependent on each other. This allows us to directly relate our method with the mathematical foundations of multi-view geometry~\cite{Hartley:2003:MVG:861369} resulting in better performance. %The details of the model are shown in Fig.~\ref{fig:camera_projection_loss} and are discussed next.%To enable the training of such a network, a large number of images and their corresponding ground truth parameters are needed. The process of generating such datasets will also be illustrated.

\textbf{Camera Model}: The camera model that we consider is the perspective projection model based on the pinhole camera~\cite{faugeras1993three}. In this work, we rely on projection of 2D image points $(u,v)$ to 3D world points $X, Y, Z$ as a reference. This leaves us with with the estimation of $13$ free parameters: focal length ($f_x$, $f_y$), principal point ($u_0$, $v_0$), disparity ($d$), baseline ($b$), pitch ($\theta_p$), translation $\mathbf{t} = \{t_x, t_y t_z\}$ and 3D coordinates $(X, Y, Z)$. %Thus, the parameters to be recovered by the network are the focal length ($f_x$, $f_y$), principal point ($u_0$, $v_0$), disparity ($d$), baseline ($b$), pitch ($\theta_p$) and translation ($t_x$, $t_y$, $t_z$).
	
%If $M$ has world coordinates $(X, Y, Z)$ and projects onto a point $m$ that has pixel coordinates $(u, v)$, the operation can be described, in homogeneous coordinates, by the equation:
%\begin{align}
% S\begin{pmatrix} u \\ v \\ 1 \end{pmatrix} = \mathbf{P}\begin{pmatrix} X \\ Y \\ Z \\ 1 \end{pmatrix}
% \label{eq:1}
% \end{align}
%  
%where $S$ is a scaling factor and the matrix $\mathbf{P}$ is in the
%format

%\begin{align}
% P = \begin{pmatrix} p_1^T & p_{14} \\ p_2^T & p_{24}\\ p_3^T & p_{34} \end{pmatrix}
% \label{eq:2}
%  \end{align}

%The $3 \times 4$ matrix $\mathbf{P}$ is commonly referred to as perspective projection %matrix and decomposed into two
%matrices: $\mathbf{P} = \mathbf{A}\mathbf{D}$ where
%
%\begin{align*}
% \mathbf{D} = \begin{pmatrix} \mathbf{R} & t \\ 0_3^T & 1\end{pmatrix} \
% \mathbf{A} = \begin{pmatrix} \alpha_u & -\alpha_u\cot\theta  & u_0 & 0\\ 0 & \frac{\alpha_v}{\sin\theta} & v_0 & 0\\0 & 0 & 1 & 0\end{pmatrix}
%\end{align*}
%
%The $4 \times 4$ matrix $\mathbf{D}$ represents the mapping from
%world coordinates to camera coordinates and accounts
%for six extrinsic parameters of the camera: three for
%the rotation $\mathbf{R}$ which is normally specified by three
%rotation (Euler) angles: $ R_x,\:R_y,\:R_z $ and three for
%the translation $ t=(t_x,\:t_y,\:t_z)^T.$ $0_3$ represents the null vector %$(0,\:0,\:0)^T.$ The $3 \times 4$ matrix $\mathbf{A}$ represents the
%intrinsic parameters of the camera: the scale factors $\alpha_u \:{and} \:\alpha_v$, the coordinates $u_0 \:{and}\: v_0$ of the principal
%point, and the angle $\theta$ between the image axes.

\textbf{Parameterization}: As revealed by previous work ~\cite{workman2015deepfocal,workman2016horizon,hold2018perceptual}, an adequate parameterization of the variables to predict can benefit convergence and final performance of the network. For the case of camera calibration, parameters such as the focal length or the tilt angles are difficult to interpret from the image content. Instead, they can be better represented by proxy parameters that are directly observable in the image. We use 2D to 3D projection as a proxy for our parameters.

%$\mathbf{A}$ can also be written as:

%\begin{align}
%\begin{pmatrix} f_x & 0 & u_0 \\ 0 & f_y & v_0 \\ 0 & 0 & 1 \end{pmatrix}
%\end{align}

A 2D point in image coordinate system is projected to camera coordinate and then to world coordinate system and the process can be explained by the following:
%
%\begin{align}
%\begin{pmatrix} u \\ v \\ 1 \end{pmatrix} \sim \begin{pmatrix} f_x & 0 & u_0 \\ 0 & f_y &%v_0 \\ 0 & 0 & 1 \end{pmatrix} \begin{pmatrix} r_{11} & r_{12} & r_{13} & t_x\\ r_{21} & r_{22} & r_{23} & t_y\\ r_{31} & r_{32} & r_{33} & t_z \end{pmatrix} \begin{pmatrix} X \\ Y \\ Z \\ 1 \end{pmatrix}
% \label{eq:3}
%  \end{align}
%

%Combining (Eq.~\ref{eq:1}) , (Eq.~\ref{eq:2}) and $\mathbf{D}$ as:

%
\begin{align}
 \begin{pmatrix} X \\ Y \\ Z \\ 1 \end{pmatrix} \sim \begin{bmatrix}\begin{pmatrix} f_x & 0 & u_0 \\ 0 & f_y & v_0 \\ 0 & 0 & 1 \end{pmatrix} \begin{pmatrix} \mathbf{R} & \mathbf{t} \\ \mathbf{0}_{3\times 1}^T & 1\end{pmatrix}\end{bmatrix}^{-1} \begin{pmatrix} u \\ v \\ 1 \end{pmatrix},
 \label{eq:4}
  \end{align}
\begin{align}
 \begin{pmatrix} X \\ Y \\ Z \\ 1 \end{pmatrix} \sim \begin{pmatrix} \mathbf{R} & \mathbf{t} \\ \mathbf{0}_{3\times 1}^T & 1\end{pmatrix}^{-1} \begin{pmatrix} f_x & 0 & u_0 \\ 0 & f_y & v_0 \\ 0 & 0 & 1 \end{pmatrix}^{-1} \begin{pmatrix} u \\ v \\ 1 \end{pmatrix}.
 \label{eq:5}
  \end{align}
where $\mathbf{R}$ is the camera rotation matrix. The image point $(u, v)$ to camera point $(x_{cam}, y_{cam}, z_{cam})$ transformation can be performed as follows (assuming skew = 0):
%
%\begin{align}
%\mathbf{A}^{-1} = \begin{pmatrix} \frac{1}{f_x} & 0 & \frac{-u_0}{f_x} \\ 0 & \frac{1}{f_y} & \frac{-v_0}{f_y} \\ 0 & 0 & 1 \end{pmatrix}
%\end{align}

%
\begin{align}
 \begin{pmatrix} y_{cam} \\ z_{cam} \\ x_{cam} \end{pmatrix} \sim \begin{pmatrix} \frac{1}{f_x} & 0 & \frac{-u_0}{f_x} \\ 0 & \frac{1}{f_y} & \frac{-v_0}{f_y} \\ 0 & 0 & 1 \end{pmatrix} \begin{pmatrix} u \\ v \\ 1 \end{pmatrix}
 \label{eq:6}
  \end{align}
\begin{subequations}
  \begin{equation}
    \label{eq-6a}
      y_{cam} = \frac{u}{f_x} - \frac{u_0}{f_x} = \frac{u - u_0}{f_x}
  \end{equation}
  \begin{equation}
    \label{eq-6b}
    z_{cam} = \frac{v}{f_y} - \frac{v_0}{f_y} = \frac{v - v_0}{f_y}
  \end{equation}
    \begin{equation}
    \label{eq-6c}
    x_{cam} = 1
  \end{equation}
\end{subequations}

Similarly, for camera to world transformation we have:

\begin{align}
 \begin{pmatrix} X \\ Y \\ Z \\ 1 \end{pmatrix} \sim \begin{pmatrix} \mathbf{R} & \mathbf{t} \\ \mathbf{0}_{3\times 1}^T & 1\end{pmatrix} \begin{pmatrix} x_{cam} \\ y_{cam} \\ z_{cam} \\ 1 \end{pmatrix}
 \label{eq:7}
  \end{align}
\begin{align}
 \begin{pmatrix} X \\ Y \\ Z \end{pmatrix} \sim \begin{pmatrix} \cos\theta & 0 & \sin\theta \\ 0 & 1 & 0 \\ -\sin\theta & 0 & \cos\theta\end{pmatrix} \begin{pmatrix} x_{cam} \\ y_{cam} \\ z_{cam} \end{pmatrix} + \begin{pmatrix} t_{x} \\ t_{y} \\ t_{z} \end{pmatrix}
 \label{eq:8}
  \end{align}
\begin{subequations}
  \begin{equation}
    \label{eq-8a}
      X = x_{cam} * \cos\theta + z_{cam} * \sin\theta + t_{x}
  \end{equation}
  \begin{equation}
    \label{eq-8b}
    Y = y_{cam} + t_{y}
  \end{equation}
    \begin{equation}
    \label{eq-8c}
    Z = -x_{cam} * \sin\theta + z_{cam} * \cos\theta + t_{z}
  \end{equation}
\end{subequations}
%

%% Finally, for camera-to-camera transformation:
%
%
%\begin{subequations}
%  \begin{equation}
%    \label{eq-9a}
%      x_W = \frac{f_x * b}{d}
%  \end{equation}
%  \begin{equation}
%    \label{eq-9b}
%    y_W = y_{cam} * x_W = -x_W * \frac{u - u_0}{f_x}
%  \end{equation}
%    \begin{equation}
%    \label{eq-9c}
%    z_W = -x_W * \frac{v - v_0}{f_y}
%\end{equation}
%\end{subequations}
%

To project a point from image to camera coordinate:
\begin{subequations}
  \begin{equation}
    \label{eq-4a}
      x_{cam} = f_x * b / d
  \end{equation}
  \begin{equation}
    \label{eq-4b}
    y_{cam} = -(x_{cam} / f_x) * (u - u_0)
  \end{equation}
    \begin{equation}
    \label{eq-4c}
    z_{cam} = (x_{cam} / f_y) * (v_0 - v)
  \end{equation}
\end{subequations}
$x_{cam}$ works as a proxy for $f_x$, baseline and disparity while $y_{cam}$ works as a proxy for $f_x$, u and $u_0$ and $z_{cam}$ works as a proxy for $f_y$, $v$ and $v_0$. Using $x_{cam}$, $y_{cam}$ and $z_{cam}$ from Eq.~\ref{eq-4a}, Eq.~\ref{eq-4b} and Eq.~\ref{eq-4c} respectively, points can be projected to world coordinate system using:
\begin{subequations}
  \begin{equation}
    \label{eq-5a}
      X = x_{cam} * \cos(\theta_p) + z_{cam} * \sin(\theta_p) + t_x
  \end{equation}
  \begin{equation}
    \label{eq-5b}
    Y = y_{cam} + t_y
  \end{equation}
    \begin{equation}
    \label{eq-5c}
    Z = -x_{cam} * \sin(\theta_p) + z_{cam} * \cos(\theta_p) + t_z
  \end{equation}
\end{subequations}
$X$ works as a proxy for pitch and $t_x$ while $Y$ works as a proxy for $t_y$ and $Z$ works as a proxy for pitch and $t_z$.

\textbf{Camera Projection Loss}: When a single architecture is trained to predict parameters with different magnitudes, special care must be taken
to weigh the loss components such that the estimation of
certain parameters do not dominate the learning process.
We notice that for the case of camera calibration, instead
of optimizing the camera parameters separately, a single
metric based on 2D to 3D projection of points can be used.

Given two images with known parameters $\omega=(f_x, f_y, u_0,\\ v_0, b, d, \theta_p, t_x, t_y, t_z, X, Y, Z)$ and a prediction of such parameters
given by the network $\hat{\omega}$=($f_x^{'}, f_y^{'}, u_0^{'}, v_0^{'}$, $b^{'}$, $d^{'}$, $\theta_p^{'}$, $t_x^{'}$, $t_y^{'}$, $t_z^{'}$, $X^{'}$, $Y^{'}$, $Z^{'}$), we get the projected point in world coordinate system through Eq.~\ref{eq-4a} - Eq.~\ref{eq-5c}. Loss is computed between actual $\omega$ and predicted  $\hat{\omega}$ using: 
  \begin{equation}
    \label{eq-6}
    L({\omega}, \hat{\omega}) = (\frac{1}{n})\sum_{i=1}^{n}MAE({\omega} , \hat{\omega})
  \end{equation}
%-------------------------------------------------------------------------

\textbf{Separating sources of loss errors}: The proposed loss solves the task balancing problem by expressing different errors in terms of a single measure. However, using several camera parameters to predict the 3D points introduces a new problem during learning: the deviation of a point from its ideal projection can be attributed to more than one parameter. In other words, an error from one parameter can backpropagate through the camera projection loss to other parameters.

To avoid this problem, we disentangle the camera projection loss, evaluating it individually for each parameter similar to~\cite{lopez2019deep}:
\begin{align*}
\begin{split}
  L_{f_x} &= L((f_x, f_y^{GT}, u_0^{GT}, v_0^{GT}, b^{GT}, d^{GT}, \theta_p^{GT},\\  &t_x^{GT}, t_y^{GT}, t_z^{GT}, X^{GT}, Y^{GT}, Z^{GT}),\omega) \\
    L_{f_y} &= L((f_x^{GT}, f_y, u_0^{GT}, v_0^{GT}, b^{GT}, d^{GT},  \theta_p^{GT},\\  &t_x^{GT}, t_y^{GT}, t_z^{GT}, X^{GT}, Y^{GT}, Z^{GT}),\omega) \\
    \ldots \\
        L_{Z} &= L((f_x^{GT}, f_y^{GT}, u_0^{GT}, v_0^{GT}, b^{GT}, d^{GT}, \theta_p^{GT},\\  &t_x^{GT}, t_y^{GT}, t_z^{GT}, X^{GT}, Y^{GT}, Z),\omega)
  \end{split}
\end{align*}
\begin{align}
\label{eq-7}
\begin{split}
  L^{*} &= \frac{L_{f_x} + L_{f_y} + L_{u_0} + ... + L_{Z}}{13}
  \end{split}
\end{align}

The loss function is further normalized to avoid the unnecessary bias due to one or more error terms by introducing weights $\alpha_i$ with each of the parameters. This bias is introduced due to heterogeneous ranges of various parameters. These weights $\alpha_i$ are learned adaptively during the training process. The updated loss function is defined as:

\begin{align}
\label{eq-9}
\begin{split}
  L^{*} &= \alpha_{f_x}L_{f_x} + \alpha_{f_y}L_{f_y} + \alpha_{u_0}L_{u_0} + ... + \alpha_{Z}L_{Z}
  \end{split}
\end{align}

%This modification of the loss function greatly increases
%convergence and final accuracy, while maintaining the main
%advantage of the camera projection loss.

\begin{figure}
    \centering
    \begin{tabular}{cc}
        {\includegraphics[width=0.35\columnwidth]{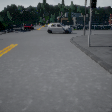}} &
        {\includegraphics[width=0.35\columnwidth]{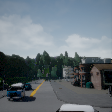}} \\
        (a) & (b) \\
        {\includegraphics[width=0.35\columnwidth]{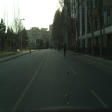}} &
        {\includegraphics[width=0.35\columnwidth]{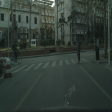}}\\
         (c) & (d)\\
     \end{tabular}
    \caption{Some representative images from the synthetic and real datasets. (a-b) CVGL (c-d) Tsinghua-Daimler.}
    \label{fig:foobar}
\end{figure}

% ---------------------------------------
\section{Results and Evaluation}

%\begin{figure}[t]
%\begin{center}
%\begin{tabular}{cc}
%	\includegraphics[ width=0.2\textwidth]{Carla/Town1/LeftRGB_00000.png}&
%	\includegraphics[ width=0.2\textwidth]{Carla/Town1/RightRGB_00000.png}\\
%	\includegraphics[ width=0.2\textwidth]{Carla/Town2/LeftRGB_00000.png}&
%	\includegraphics[ width=0.2\textwidth]{Carla/Town2/RightRGB_00000.png}
%	%\fbox{\includegraphics[width=1.5cm]{images/stadium_Sat.jpg}}\\
	%Freeway &Mountain &Palace &River &Ship &Stadium
%\end{tabular}
%\end{center}
%\caption{Example stereo image pairs from the generated dataset.}
%\label{fig:Dataset}
%\end{figure}

%\begin{figure}[t]
%\begin{center}
%\begin{tabular}{cc}
%	\includegraphics[width=0.45\columnwidth]{images/Town01.jpg}&
%	\includegraphics[width=0.45\columnwidth]{images/Town02.jpg}\\
%\end{tabular}
%\end{center}
%\caption{Street maps of towns from CARLA~\cite{Dosovitskiy17} Simulator (a) Town 1. (b) Town 2.}
%\label{fig:town}
%\end{figure}

\begin{table*}[t]

\caption{Table showing Normalized MAE in predicted parameters on CVGL test set comprising of 36,905 images.}
\centering
%\begin{tabular}{ | m{2cm} | m{2cm}| } 
%\resizebox{0.99\columnwidth}{!}
{
\begin{adjustbox}{max width=\textwidth}
\begin{tabular}{ l|c|c|c|c|c|c|c|c|c|c } 
%\multirow{2}{*}{} &  \multicolumn{6}{c}{\textbf{MAE}}\\
%\cline{2-7}
\hline
 & $f_x$ & $f_y$ & $u_0$ & $v_0$ & $b$ & $d$ & $t_x$ & $t_y$ & $t_z$ & $\theta_p$\\
 \hline
Average~\cite{workman2015deepfocal} & 1.003 & 1.539 & 1.326 & 1.200 & \textbf{-1.908} & \textbf{-6.624} & -0.562 & 94.233 & \textbf{-14.031} & \textbf{-3.123}\\
Deep-Homo~\cite{detone2016deep} & 0.055 & 0.055 & 0.018 & 0.018 & -0.091 & -0.167 & -0.091 & 1.900 & -1.553 & -0.258\\
MTL-CPL-U (Ours) & 0.713 & 0.573 & 0.854 & 0.790 & -0.310 & -1.447 & \textbf{-0.722} & 3.085 & -1.165 & -1.392\\
MTL-Baseline (Ours)  & 0.387 & 0.445 & 0.311 & 0.313 & -0.240 & -0.449 & -0.258 & 0.170 & -0.424 & -0.405\\
MTL-CPL-A (Ours) & \textbf{0.005} & \textbf{0.003} & \textbf{0.007} & \textbf{0.011} & -0.013 & -0.121 & -0.013 & \textbf{0.093} & -0.050 & -0.024\\
\hline
\end{tabular}
\end{adjustbox}
\label{table:1}
}
\end{table*}

\begin{table*}[t]

\caption{Table showing Normalized MAE in predicted parameters on Tsinghua-Daimler test set comprising of 2,914 images. For this experiment, we just did a forward pass without any transfer learning or training.}
\centering
%\begin{tabular}{ | m{2cm} | m{2cm}| } 
%\resizebox{0.99\columnwidth}{!}
{
\begin{adjustbox}{max width=\textwidth}
\begin{tabular}{ l|c|c|c|c|c|c|c|c|c|c } 
%\multirow{2}{*}{} &  \multicolumn{6}{c}{\textbf{MAE}}\\
%\cline{2-7}
\hline
 & $f_x$ & $f_y$ & $u_0$ & $v_0$ & $b$ & $d$ & $t_x$ & $t_y$ & $t_z$ & $\theta_p$\\
 \hline
Average~\cite{workman2015deepfocal} & 1.000 & 1.014 & 1.017 & 1.021 & 364.633 & 4.893 & 31.167 & 327.128 & 5.109 & 2321.338\\
Deep-Homo~\cite{detone2016deep} & 0.971 & 0.971 & 0.947 & 0.896 & 350.476 & 1.538 & 37.494 & \textbf{10.470} & 2.182 & \textbf{193.760}\\
MTL-CPL-U (Ours) & \textbf{0.950} & \textbf{0.944} & \textbf{0.896} & \textbf{0.869} & \textbf{173.574} & 2.634 & 81.817 & 44.099 & \textbf{1.497} & 290.614\\
MTL-Baseline (Ours)  & 0.958 & 0.958 & 0.946 & 0.894 & 237.554 & 1.563 & \textbf{26.660} & 27.387 & 1.637 & 354.791\\
MTL-CPL-A (Ours) & 0.952 & 0.956 & 0.946 & 0.895 & 363.883 & \textbf{1.515} & 44.958 & 33.421 & 1.785 & 324.892\\
\hline
\end{tabular}
\end{adjustbox}
\label{table:2}
}
\end{table*}

%\begin{table}[b]

%\caption{Table showing percentage of correct results at different FOV error threshold in %degrees.}
%\centering
%%\begin{tabular}{ | m{2cm} | m{2cm}| } 
%\resizebox{.99\columnwidth}{!}
%{
%\begin{tabular}{ l|c|c|c|c|c|c } 
%\multirow{2}{*}{} &  \multicolumn{6}{c}{\textbf{Correct (\%)}}\\
%\cline{2-7}
% & 0 & 1 & 2 & 3 & 4 & 5\\
%\hline
%Average~\cite{workman2015deepfocal} & 0.00 & 56.59 & 100.00 & 100.00 & 100.00 & 100.00\\
%Deep-Homo~\cite{detone2016deep} & 0.00 & 94.60 & 100.00 & 100.00 & 100.00 & 100.00\\
%MTL-CPL-U (Ours) & 0.00 & 96.54 & 99.92 & 100.00 & 100.00 & 100.00\\
%MTL-Baseline (Ours) & 0.00 & 95.95 & 99.85 & 100.00 & 100.00 & 100.00\\
%MTL-CPL-A (Ours) & 0.00 & \textbf{99.78} & 99.99 & 100.00 & 100.00 & 100.00\\
%\hline
%\end{tabular}
%\label{table:2}
%}
%\end{table}

%-------------------------------------------------------------------------
\subsection{Datasets}

\textbf{Synthetic Data}: We trained and evaluated our proposed approach by generating a new CVGL Camera Calibration dataset using Town 1 and Town 2 of CARLA~\cite{Dosovitskiy17} Simulator. %Sample images along with town maps are shown in Fig.~\ref{fig:Dataset} and Fig.~\ref{fig:town} respectively. 
The dataset consists of $50$ camera configurations with each town having $25$ configurations. The parameters modified for generating the configurations include  $fov$, $x$, $y$, $z$, pitch, yaw, and roll. Here, $fov$ is the field of view, (x, y, z) is the translation while (pitch, yaw, and roll) is the rotation between the cameras. The total number of image pairs is $1,23,017$, out of which $58,596$ belong to Town 1 while $64,421$ belong to Town 2, the difference in the number of images is due to the length of the tracks.

\textbf{Real Data}: We have used a recent Cyclist Detection dataset~\cite{li2016new} for evaluating our approach on real world data. %to get better understanding of the working model on unseen scenarios. 
We have used the test set provided by the authors containing $2,914$ images by first deriving the right image using left and disparity images and then use the pair as input to compare different methods.

%In this paper, we investigate a new loss function for automatic calibration from a single image. To train the deep learning CNN, we need numerous training examples. 

%However, to the best of our knowledge, there is no existing large-scale dataset of images with ground truth intrinsic and extrinsic parameters that could be used to train a deep learning network. 

%Concretely, to train our network, we need millions of images with different focal lengths, along with the corresponding ground truth values. In practice, it is cumbersome and virtually impossible to manually capture such data.

%Furthermore, to the best of our knowledge the SUN360 database~\cite{xiao2012recognizing} used in~\cite{zhang2020deepptz,bogdan2018deepcalib,lopez2019deep} is no more publicly available. 

\textbf{Implementation Details}: Our loss is implemented and trained using Keras~\cite{ketkar2017introduction}, an open-source deep learning framework. All networks are trained on GeForce GTX $1050$ Ti GPU for $200$ epochs with early stopping using ADAM optimizer~\cite{KingmaADAM2015} with Mean Absolute Error (MAE) loss function and a base learning rate $\eta$ of ${10}^{-3}$ with a batch size of $16$.

The Lambda layer exists in Keras so that arbitrary expressions can be used as a Layer when constructing Sequential and Functional API models. In the proposed architecture, Lambda layers have been utilized for basic operations including addition, subtraction, multiplication, division, negation, cosine and sine. The intuition is to incorporate mathematical equations in the learning framework.

We have used Normalized Mean Absolute Error for evaluation as follows:

  \begin{equation}
    \label{eq-60}
    NMAE(y, \hat{y}) = \frac{MAE(y, \hat{y})}{\frac{1}{n}\sum_{i=1}^{n}|y_{i}|} = \frac{MAE(y, \hat{y})}{mean(|y|)}
  \end{equation}
where $y$ and $\hat{y}$ are the target and estimated values respectively.

\subsection{Comparative Analysis}

\textbf{Experimental Setup}: We compared our proposed method with two state-of-the-art approaches namely Average field of view~\cite{workman2015deepfocal} and Deep-Homo~\cite{detone2016deep}. Average field of view~\cite{workman2015deepfocal} is a baseline approach, given a query image, it uses the average field of view of the training set as the prediction~\cite{workman2015deepfocal}. {Deep-Homo}~\cite{detone2016deep} estimates an 8-degree-of-freedom homography between two images. We have modified Deep-Homo\cite{detone2016deep} to predict the required 13 parameters for comparison purposes as by default, it only predicted 8 values corresponding to the four corners and then using 4-point parameterization and then convert it into the homography matrix. For the purpose of the ablative study, we also created three variants of our multi-task learning approach namely MTL-Baseline, MTL-CPL-U, and MTL-CPL-A. MTL-Baseline does not include any additional layers to incorporate camera model equations, instead, it is an end-to-end learning architecture based on mean absolute error (MAE). It has 13 regressors sharing a common feature extractor directly predicting the required values. MTL-Baseline is implemented to study the effect of proposed camera projection loss. We also used two variants of camera projection loss one with uniform weighting (MTL-CPL-U) in the loss function and the other with adaptive weighting (MTL-CPL-A) to balance the heterogeneous ranges of calibration parameters. 

%\subsubsection{Deep-Homo}

%We have modified Deep-Homo\cite{detone2016deep} to predict the required 13 parameters for %comparison purposes as by default, it only predicted 8 values corresponding to the four %corners and then using 4-point parameterization and then convert it into the homography %matrix using the function
%getPerspectiveTransform()in OpenCV.

\textbf{Error Analysis on generated data}: We compare the normalized mean absolute error (NMAE) of each of the parameters by all the methods with our proposed approach. It can be seen from Table~\ref{table:1} that for focal length ($f_x$,$f_y$), principal point offset ($u_0$,$v_0$)  and translation in y-axis ($t_y$) MTL-CPL-A  approach resulted in minimum values for NMAE while for translation  in  x-axis ($t_x$) MTL-CPL-U performs  better while for all other parameters Average performs better due to bias in loss introduced as a result of the heterogeneous range of values among parameters. This indicates that incorporating camera model geometry in the learning framework not only resulted in a more interpretable learning framework but it also outperforms the state-of-the-art methods.

\textbf{Error Analysis on real data}: For this experiment, we didn't trained on the Tsinghua-Daimler dataset but just performed a forward pass to test the generalizability and the results further strengthen our argument. It can be seen from Table~\ref{table:2} that our proposed multi-task learning approach, MTL-CPL-U, outperforms other methods on $6$ out of $10$ parameters. For pitch ($\theta_p$) and translation in y-axis ($t_y$) Deep-Homo performs better due to bias in loss introduced as a result of the heterogeneous range of values among parameters. For all the remaining parameters our proposed multi-task learning approach resulted in minimum values for NMAE which further solidifies  our argument of incorporating camera model geometry in the learning framework.

\section{Conclusion}
Our proposed approach outperforms several baselines, including CNN-based methods on both synthetic and real data is an amalgam of Deep Learning and closed-form analytical solutions. %This inclusion of mathematical equations within a CNN offers better interpretation along with superior performance. 
Thus this paper offers a significantly new idea in the area of Machine Learning and Computer Vision. Although we have applied the proposed solution for the estimation of camera calibration parameters, it offers a general framework for many such problems. For example,  tracking using Kalman Filter, homography estimation, and many other applications we plan to implement in future.

%\vfill\pagebreak

\bibliographystyle{IEEEbib}
\bibliography{strings,refs}

\begin{thebibliography}{10}

\bibitem{Dosovitskiy17}
Alexey Dosovitskiy, German Ros, Felipe Codevilla, Antonio Lopez, and Vladlen
  Koltun,
\newblock ``{CARLA}: {An} open urban driving simulator,''
\newblock in {\em 1st Annual Conference on Robot Learning}, 2017, pp. 1--16.

\bibitem{szegedy2016rethinking}
Christian Szegedy, Vincent Vanhoucke, Sergey Ioffe, Jon Shlens, and Zbigniew
  Wojna,
\newblock ``Rethinking the inception architecture for computer vision,''
\newblock in {\em IEEE Conference on Computer Vision and Pattern Recognition},
  2016, pp. 2818--2826.

\bibitem{workman2015deepfocal}
Scott Workman, Connor Greenwell, Menghua Zhai, Ryan Baltenberger, and Nathan
  Jacobs,
\newblock ``Deepfocal: A method for direct focal length estimation,''
\newblock in {\em IEEE International Conference on Image Processing}, 2015, pp.
  1369--1373.

\bibitem{rong2016radial}
Jiangpeng Rong, Shiyao Huang, Zeyu Shang, and Xianghua Ying,
\newblock ``Radial lens distortion correction using convolutional neural
  networks trained with synthesized images,''
\newblock in {\em Asian Conference on Computer Vision}. Springer, 2016, pp.
  35--49.

\bibitem{hold2018perceptual}
Yannick Hold-Geoffroy, Kalyan Sunkavalli, Jonathan Eisenmann, Matthew Fisher,
  Emiliano Gambaretto, Sunil Hadap, and Jean-Fran{\c{c}}ois Lalonde,
\newblock ``A perceptual measure for deep single image camera calibration,''
\newblock in {\em IEEE Conference on Computer Vision and Pattern Recognition},
  2018, pp. 2354--2363.

\bibitem{lopez2019deep}
Manuel Lopez, Roger Mari, Pau Gargallo, Yubin Kuang, Javier Gonzalez-Jimenez,
  and Gloria Haro,
\newblock ``Deep single image camera calibration with radial distortion,''
\newblock in {\em IEEE Conference on Computer Vision and Pattern Recognition},
  2019, pp. 11817--11825.

\bibitem{zhai2016detecting}
Menghua Zhai, Scott Workman, and Nathan Jacobs,
\newblock ``Detecting vanishing points using global image context in a
  non-manhattan world,''
\newblock in {\em IEEE Conference on Computer Vision and Pattern Recognition},
  2016, pp. 5657--5665.

\bibitem{detone2016deep}
Daniel DeTone, Tomasz Malisiewicz, and Andrew Rabinovich,
\newblock ``Deep image homography estimation,''
\newblock {\em arXiv preprint arXiv:1606.03798}, 2016.

\bibitem{bogdan2018deepcalib}
Oleksandr Bogdan, Viktor Eckstein, Francois Rameau, and Jean-Charles Bazin,
\newblock ``Deep{C}alib: a deep learning approach for automatic intrinsic
  calibration of wide field-of-view cameras,''
\newblock in {\em ACM SIGGRAPH European Conference on Visual Media Production},
  2018, pp. 1--10.

\bibitem{zhang2020deepptz}
Chaoning Zhang, Francois Rameau, Junsik Kim, Dawit~Mureja Argaw, Jean-Charles
  Bazin, and In~So Kweon,
\newblock ``Deep{PTZ}: Deep self-calibration for ptz cameras,''
\newblock in {\em IEEE/CVF Winter Conference on Applications of Computer
  Vision}, 2020, pp. 1041--1049.

\bibitem{workman2016horizon}
Scott Workman, Menghua Zhai, and Nathan Jacobs,
\newblock ``Horizon lines in the wild,''
\newblock {\em arXiv preprint arXiv:1604.02129}, 2016.

\bibitem{barreto2006unifying}
Jo{\~a}o~P Barreto,
\newblock ``A unifying geometric representation for central projection
  systems,''
\newblock {\em Computer Vision and Image Understanding}, vol. 103, no. 3, pp.
  208--217, 2006.

\bibitem{russakovsky2015imagenet}
Olga Russakovsky, Jia Deng, Hao Su, Jonathan Krause, Sanjeev Satheesh, Sean Ma,
  Zhiheng Huang, Andrej Karpathy, Aditya Khosla, Michael Bernstein, et~al.,
\newblock ``Imagenet large scale visual recognition challenge,''
\newblock {\em International Journal of Computer Vision}, vol. 115, no. 3, pp.
  211--252, 2015.

\bibitem{Hartley:2003:MVG:861369}
Richard Hartley and Andrew Zisserman,
\newblock {\em Multiple View Geometry in Computer Vision},
\newblock Cambridge University Press, New York, NY, USA, 2 edition, 2003.

\bibitem{faugeras1993three}
Olivier Faugeras and Olivier~Autor Faugeras,
\newblock {\em Three-Dimensional Computer Vision: A Geometric Viewpoint},
\newblock MIT press, 1993.

\bibitem{li2016new}
Xiaofei Li, Fabian Flohr, Yue Yang, Hui Xiong, Markus Braun, Shuyue Pan,
  Keqiang Li, and Dariu~M Gavrila,
\newblock ``A new benchmark for vision-based cyclist detection,''
\newblock in {\em IEEE Intelligent Vehicles Symposium}, 2016, pp. 1028--1033.

\bibitem{ketkar2017introduction}
Nikhil Ketkar,
\newblock ``Introduction to {K}eras,''
\newblock in {\em Deep learning with Python}, pp. 97--111. Springer, 2017.

\bibitem{KingmaADAM2015}
Diederik~P. Kingma and Jimmy Ba,
\newblock ``Adam: {A} method for stochastic optimization,''
\newblock in {\em International Conference on Learning Representations}, San
  Diego, CA, USA, 2015.

\end{thebibliography}

\end{document}